\documentclass{ecai} 

\usepackage{latexsym}
\usepackage{amssymb}
\usepackage{amsmath}
\usepackage{amsthm}
\usepackage{booktabs}
\usepackage{enumitem}
\usepackage{graphicx}
\usepackage{color}
\usepackage{algorithm}
\usepackage{algpseudocode}

\newcommand{\BibTeX}{B\kern-.05em{\sc i\kern-.025em b}\kern-.08em\TeX}

\begin{document}


\begin{frontmatter}


\paperid{1210} 


\title{FAIRGAME: a Framework for AI Agents Bias Recognition using Game Theory} 



\author[A]{\fnms{Alessio}~\snm{Buscemi}\orcid{0009-0003-4668-9915}\thanks{Corresponding Author. Email:  alessio.buscemi@list.lu}\footnote{Equal contribution.}}
\author[B]{\fnms{Daniele}~\snm{Proverbio}\orcid{0000-0002-0122-479X}\footnotemark}
\author[C]{\fnms{Alessandro}~\snm{Di Stefano}\orcid{0000-0003-4905-3309}} 
\author[C]{\fnms{The Anh}~\snm{Han}\orcid{0000-0002-3095-7714}} 
\author[A]{\fnms{German}~\snm{Castignani}\orcid{0000-0001-5594-4904}} 
\author[D]{\fnms{Pietro}~\snm{Liò}\orcid{0000-0002-0540-5053}} 

\address[A]{Luxembourg Institute of Science and Technology}
\address[B]{Department of Industrial Engineering, University of Trento}
\address[C]{School Computing, Engineering and Digital Technologies, Teesside University}
\address[D]{Department of Computer Science and Technology, University of Cambridge}


\begin{abstract}
Letting AI agents interact in multi-agent applications adds a layer of complexity to the interpretability and prediction of AI outcomes, with profound implications for their trustworthy adoption in research and society. Game theory offers powerful models to capture and interpret strategic interaction among agents, but requires the support of reproducible, standardized and user-friendly IT frameworks to enable comparison and interpretation of results. To this end, we present FAIRGAME, a Framework for AI Agents Bias Recognition using Game Theory. We describe its implementation and usage, and we employ it to uncover biased outcomes in popular games among AI agents, depending on the employed Large Language Model (LLM) and used language, as well as on the personality trait or strategic knowledge of the agents. Overall, FAIRGAME allows users to reliably and easily simulate their desired games and scenarios and compare the results across simulation campaigns and with game-theoretic predictions, enabling the systematic discovery of biases, the anticipation of emerging behavior out of strategic interplays, and empowering further research into strategic decision-making using LLM agents.
\end{abstract}

\end{frontmatter}


\section{Introduction}

AI agents powered by Large Language Models (LLM) are increasingly used in research \cite{lu2024llms}, social \cite{tessler2024ai} and industrial applications \cite{patel2020leveraging,stone2020artificial}, calling for the development of accurate prediction frameworks for their behaviors during interactions among themselves or with humans. Reliable predictions are essential for developing novel applications, promoting trustworthy AI systems, and mitigating undesirable outcomes and biases \cite{buscemi2024chatgpt, fulgu2024surprising}. Numerous approaches have been developed to improve the transparency and interpretability of individual AI agents \cite{ali2025entropy, el2025towards}, as well as to identify their inconsistencies, biases and hallucinations \cite{buscemi2024newspapers, li2023halueval,li2023evaluating}. However, less is known about cases where multiple interacting agents are involved \cite{hammond2025multi}, where emerging biases may skew strategic outputs in unpredictable manners. Studies emulating human behaviors \cite{park2023generative} may produce spurious predictions. Also, in industry applications such as automated dispute resolution \cite{brooks2022artificial, falcao2024making}, auction design or pricing mechanisms in finance and economics \cite{ bahtizin2019using, chaffer2025governing, chen2023utility}, or supply chain negotiations \cite{abaku2024theoretical, min2010artificial, ramachandran2022contract}, hidden biases may result in unjust decisions, disproportionate favoring of certain groups, and distortion of fair competition. 

To address multi-agent interactions, and in addition to methods tailored to individual agents, game theory \cite{owen2013game} formalizes interactions as games, so as to model, predict and optimize the strategic responses of rational agents \cite{balabanova2025media, falcao2024making, HanAICom2022emergent}. Game theory has been employed to model and understand human choices in various contexts \cite{stewart2024distorting, talajic2024strategic}, and AI-based players have been increasingly tested to reproduce classical game scenarios and provide additional complexity to them \cite{fontana2024nicer, wang2024large, willis2025will}, as well as to interact in game-like distributed technologies \cite{he2025generative}. However, due to varying research protocols and discipline constraints, bridging the gap between theoretical game theory and empirical investigations of LLM agents in a reproducible, systematic and user-friendly setting is still a challenge \cite{mao2023alympics}.

To facilitate seamless and reproducible integration of game-theoretic evaluations of AI interactions, we introduce FAIRGAME (Framework for AI Agents Bias Recognition using Game Theory), a versatile framework designed to simulate diverse scenarios, ranging from classical game theory models to realistic industrial use cases. FAIRGAME is an open-source project \cite{githubFairgame}. 

In FAIRGAME, AI agents can be equipped with distinct features, such as strategic attitudes and personalities, linguistic variations, cultural orientations and more. The framework allows  quantitative simulations of arbitrarily complex games in a systematic and reproducible manner, and to observe the emerging outcomes brought about by strategic interactions \cite{han2021or, buscemi2025llms}, enabling direct comparison with game-theoretic predictions and supporting the inference of strategies through observations of multiple experiments \cite{montero2022inferring}. Incorporating AI agents into controlled and predictable scenarios will also help identify and mitigate hidden biases related to language, cultural attributes, and more, which could result in suboptimal outcomes, unjust advantages, ethical dilemmas, or systemic inefficiencies \cite{cabrera2023ethical, ferrara2023fairness, gichoya2023ai}.

In this paper, we present the implementation of FAIRGAME, and evaluate its usage and outputs across multiple games, human languages, and LLMs. Through several use cases, we show that our empirical results recognize LLM biases in strategic interactions and identify previously unknown inconsistencies across the LLMs used to develop the agents. Overall, our results suggest that AI agents may exhibit suboptimal behaviors when interacting through different languages and game contexts, and may deviate even significantly from game-theoretic predictions. This supports the use of reproducible and controlled simulation pipelines to predict the interacting behavior of LLM agents, which defy classical modeling approaches. Finally, we propose a scoring system to evaluate and compare the sensitivity of LLMs to game determinants and strategies, so as to guide the selection of LLMs for the development of game-theoretic experiments and strategic AI applications. 

In the following, we first provide a comprehensive overview of FAIRGAME, detailing its implementation and operational usage. Then, we present our use cases: two common games in different variants and languages, with agents instantiated on different LLMs and equipped with varying personalities and degrees of knowledge about the game progression. Finally, we present our experimental findings and scoring system, showing FAIRGAME’s efficacy in detecting biases and inconsistencies in AI-game-theoretic analyses.


\section{Methodology: introducing FAIRGAME}
\label{sec:methodology}

\begin{figure}[t]
    \centering
    \includegraphics[width=0.85\linewidth]{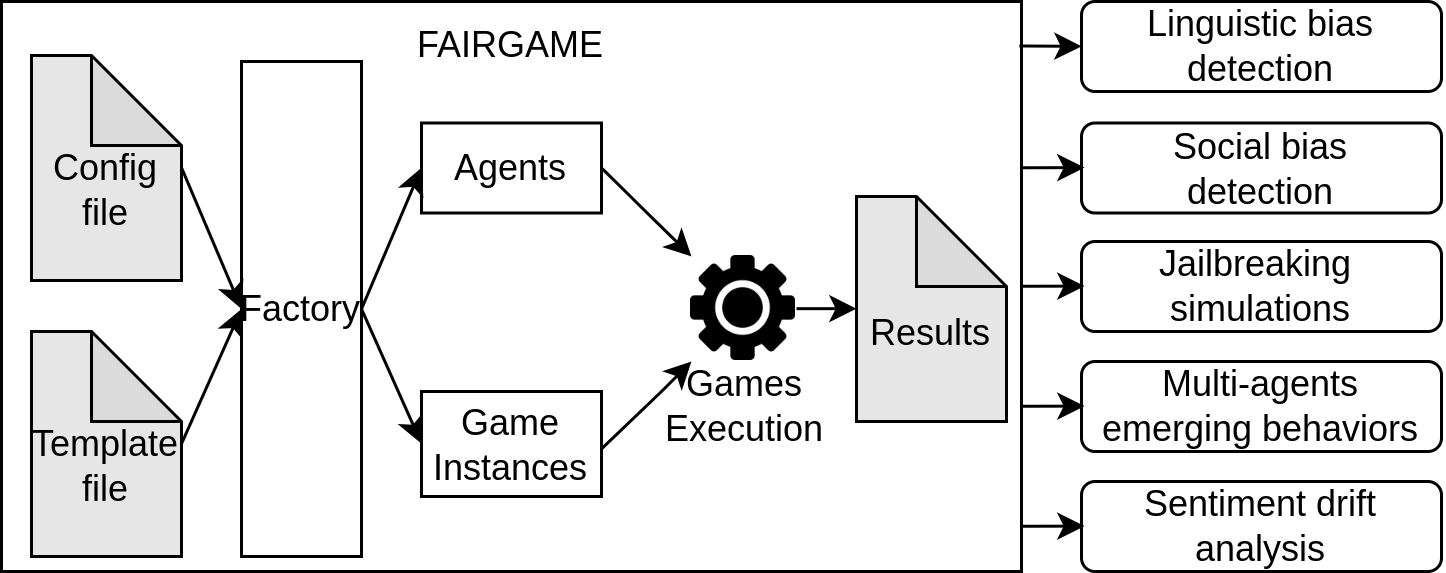}
    \vspace{-2mm}
    \caption{Schematic representation of FAIRGAME flow of document dependencies and outputs.
    }
    \label{fig:visual_abstract}
\end{figure}

FAIRGAME is a computational framework interfacing user-defined instructions to create the desired agents, eventually delivering standardized outputs for subsequent processing (Fig. \ref{fig:visual_abstract}). It allows to test user-defined games, described textually via prompt injection and including any desired payoff matrix, as well as to define traits of the agents. The agents can be built from any LLM of choice, by calling the corresponding APIs (several of which are already provided in our package; however, any usage fees are covered by the end user). 
FAIRGAME requires the following inputs. First, a \textbf{Configuration File:} a JSON file that defines the setup of both the agents and the game, in terms of parameters, payoff matrix entries, and additional information for the agents. For instance, agents can be associated with certain personalities, so as to increase the complexity of human behavior emulators \cite{ he2024afspp, newsham2025inducing} and predict the responses of personality-driven autonomous agents \cite{klinkert2024driving, newsham2025personality}. Table \ref{tab:game_config} provides a detailed overview of the fields of the configuration file, along with their explanation. 
An example of configuration file is in Supplementary Material Section S1 \cite{suppFairgame}.
A\textbf{Prompt Template:} a text file that defines the instruction template, providing a literal description of the game. It includes the instructions for each agent, with each round's parameters filled in from the configuration file, allowing customization of each agent. At runtime, this template is transformed into a prompt that includes details on available strategies, the corresponding payoffs based on decisions, and other configuration-specified factors -- such as an agent’s personality and awareness of the opponent's personality -- as well as prior history information in case of repeated games. This file can be translated in any language of choice, allowing for multilingual tests. 
An example of an English prompt template for a Prisoner’s Dilemma game is provided in Supplementary Material Section S2 \cite{suppFairgame}. 
More than one prompt template can be associated to the same \texttt{config} file.

\begin{table*}[htbp]
\scriptsize
\centering
\caption{Game Configuration Fields and Their Explanations}
\begin{tabular}{lp{3cm}p{2cm}p{9cm}}
\hline
\textbf{Field}  & \textbf{Subfield} & \textbf{Type} & \textbf{Explanation}\\
\hline
name & & String & Represents the name of the game or scenario being simulated.\\
nRounds & & Integer & Specifies the maximum number of rounds to be played in the game.\\
nRoundsIsKnown & & Boolean & Indicates whether the agents know the maximum number of rounds (True if known, False otherwise).\\
llm & & String & Defines which LLM will be used to simulate the agents.\\
languages & & List & Lists the human languages in which the agents can be queried.\\
allAgentPermutations & & Boolean & Specifies whether to compute all permutations of agent configurations or create only one instance of the agents based on provided configurations.\\
agents &  & Dictionary & Contains configurations of the agents.\\
       & names & List of Integers & Identifiers or names of the agents.\\
       & personalities & Dictionary & Defines agents' personalities, translated for each language. 
       If a personality is 'None', it means that this information will be omitted from the prompt. \\
       & opponentPersonalityProb & List of Integers & Probability that a certain agent has a certain personality, as referred to the other agents.
       If the probability is 0, this information is omitted from the prompt.\\
payoffMatrix &  & Dictionary & Contains information about the game's payoff matrix.\\
               & weights & Dictionary & Specifies the weight values used in the payoff matrix.\\
               & strategies & Dictionary& Details strategies agents can adopt, translated for each language in languages.\\
               & combinations & Dictionary & Enumerates the possible combinations of strategies that both agents choose in each round.\\
               & matrix & Dictionary & Maps each combination to weights (payoffs) that agents receive when such scenarios occur.\\
stopGameWhen & & List of Strings & Specifies combinations in the payoff matrix that trigger the end of the game during a round.\\
agentsCommunicate & & Boolean & If True, the agents exchange a message with each other at each round before making a decision; if False, they do not communicate.\\
\hline
\end{tabular}
\label{tab:game_config}
\end{table*}

\subsection{Creation and execution of games}
\label{sub:game_creation}

Algorithm \ref{alg1} processes the configuration file \textit{CF} and the set of prompt templates \( PT \) as inputs, and outputs a list \( G \) containing all instantiated games. The process involves validating inputs, extracting relevant game information, configuring agents, and creating the games.

First, the configuration file \textit{CF} is validated to ensure that it conforms to the required structure and contains all necessary information (step 1). Similarly, the prompt templates \textit{PT} are validated against \textit{CF} to confirm compatibility and completeness (step 2). After that, key information regarding the game to create is extracted from the configuration file, including the list of languages (\textit{langs}) and the LLM (\textit{llm}) and whether to compute all agent configuration permutations (\textit{all\_agent\_perm}) (steps 3). 
If \texttt{true}, the function \textit{all\_agent\_perm} generates all possible agent configurations of the personalities and probabilities of knowing the opponent personality. Otherwise, pre-defined agent configurations are retrieved using the \textit{get\_agents\_config} function (step 4-8). An empty list \( G \) is then initialized to store the created games (step 9). The algorithm iterates through each agent configuration (\textit{agent\_config}) to create the games. For each configuration, agents are instantiated using the \textit{create\_agents} function, which sets up agents according to their configurations and the selected language model \textit{llm} (step 10). A game instance is then created using \textit{create\_game}, which takes the information about the game, the agent details, and prompt templates as inputs (step 11). The created game is appended to the list \( G \) (step 12).
At the end of the process, \( G \) contains all the instantiated games, each configured with the appropriate agents, game parameters, and rules, in suitable format for the LLMs.

\begin{algorithm}[b]
\scriptsize
	\begin{algorithmic}[1]
	\Require $CF$: Configuration file, $PT$: Prompt templates
	\Ensure $G$: list of all instantiated games
    \State validate\_config\_file($CF$)
    \State validate\_templates($PT$, $CF$)
    \State \textit{game\_info}, \textit{langs}, \textit{llm}, \textit{all\_agent\_perm} $\gets$ extract\_info($CF$) 
    \If {\textit{all\_agent\_perm}}
    \State \textit{agents\_config} $\gets$ compute\_agents\_combos($CF$, \textit{langs})
    \Else
    \State \textit{agents\_config} $\gets$ get\_agents\_config($CF$, \textit{langs})
    \EndIf
    \State $G$ $\gets$ ()
    \For {$ac$ \textbf{in} \textit{agents\_config}}
        \State \textit{agents} $\gets$ create\_agents($ac$, \textit{llm})
        \State \textit{game} $\gets$ create\_games(\textit{game\_info}, \textit{agents}, $PT$)
        \State $G$.append(\textit{game})
    \EndFor
    \end{algorithmic}
 \caption{Creation of games}
    \label{alg1}
\end{algorithm}


Once instantiated, games are executed as per Algorithm \ref{alg2}. An empty list \( O \) is initialized (step 1), which will be populated with the outcomes of all games after execution. The list \( G \) is then given as input to the algorithm. Each game \( g \) in \( G \) is processed independently (steps 2–9). If the user requests multiple rounds in an evolutionary game theory setting, a \texttt{while} loop governs their execution. The loop continues as long as two conditions are satisfied: the current round count does not exceed the maximum number of rounds \( g.\textit{n\_rounds} \), and the game-specific stopping condition \( g.\textit{stop\_cond} \) is not met (step 4). The stopping condition allows for the game to terminate early based on predefined criteria, ensuring flexibility in simulation.
Within each iteration of the \texttt{while} loop, the function \( g.\textit{run\_round()} \) is called to execute the logic for the current round (step 5). 
This function, which is described in detail in Algorithm \ref{alg3}, simulates the actions of the agents involved in the game. Note that communication between agents is supported by FAIRGAME; however, we do not use it in the next use cases, hence the algorithm described here represents a simplified version that does not take inter-agent communication into account.
We have dedicated a separate study to analyzing the effect of communication by comparing outcomes with and without it \cite{buscemi2025strategic}.

Once the game terminates -— either because the maximum number of rounds has been reached or the stopping condition is satisfied -— the algorithm retrieves the game's history, which is appended to the output file \( O \) (step 8). 

\begin{algorithm}[t]
\scriptsize
	\begin{algorithmic}[1]
	\Require $G$: list of all instantiated games
	\Ensure $O$: list of outcomes of all games
        \State $O$ $\gets$ ()
        \For {$g$ \textbf{in} $G$}
        \State \textit{round} $\gets$ 1
        \While {\textit{round} $\leq$ $g$.\textit{n\_rounds} \textbf{and} not\_met($g$.\textit{stop\_cond})}
        \State $g$.run\_round()
        \State \textit{round}++
        \EndWhile
         \State $O$.append($g$.history())
        \EndFor
    \end{algorithmic}
 \caption{Execution of games}
    \label{alg2}
\end{algorithm}

Algorithm \ref{alg3} describes how actions are simulated.
For each agent, the algorithm first retrieves its opponent agent -- or agents, in case of more than two players (step 2). It then determines the appropriate template language for the agent, which is based on its language preference (step 3). Using this template language, a prompt is created that incorporates key elements of the game's current state, such as the total number of rounds, the current round, whether the number of rounds is known, the payoff matrix, and the game's history (step 4).
Next, the agent chooses a strategy for the current round based on the generated prompt (step 5). The corresponding payoff for this strategy is then computed (step 6), and the game's history is updated with the agent's move and resulting payoff (step 7).
After both agents completed their actions, FAIRGAME proceeds to the next round, and the process continues until all rounds are executed, as per Algorithm \ref{alg2}.

\begin{algorithm}[t]
\scriptsize
	\begin{algorithmic}[1]
	\Require $g$: game
	\Ensure The history of the game is updated
        \For {\textit{agent} \textbf{in} \textit{agents}}
        \State \textit{opponents} $\gets$ get\_opponents(\textit{agents})
        \State \textit{template\_lang} $\gets$ get\_template($g$.\textit{templates}, \textit{agent}.\textit{lang})
        \State \textit{prompt} $\gets$ 
        create\_prompt(\textit{template\_lang}, $g$.\textit{n\_rounds}, $g$.\textit{current\_round}, $g$.\textit{n\_rounds\_known}, $g$.\textit{payoff\_matrix}, $g$.history()) 
        \State \textit{strategy} $\gets$ \textit{agent}.choose\_strategy\_round(\textit{prompt}) 
        \State \textit{payoff} $\gets$ compute\_payoff(\textit{strategy})
        \State $g$.update\_history(\textit{agent}, \textit{payoff})
        \EndFor
    \end{algorithmic}
 \caption{Run round}
    \label{alg3}
\end{algorithm}

\section{LLM-based game-theoretic experiments}
Prior research \cite{balabanova2025media, fontana2024nicer, wang2024large} demonstrates that LLMs do not always comply with predictions from game theory. Instead, they exhibited consistently cooperative behavior when engaging in traditional game-theoretic scenarios. To systematically investigate the emergence of strategic behaviors, we employ FAIRGAME on a set of games, languages, LLMs and agent features, as described below.

\begin{table*}[h!]
\scriptsize
    \centering
    \begin{tabular}{|p{2cm}|p{1.9cm}|p{1.6cm}|p{1.8cm}|p{1.6cm}|p{2.7cm}|p{3.5cm}|} 
        \hline
        \textbf{Model} & \textbf{Provider} & \textbf{No. Params.} & \textbf{Licensing Type} & \textbf{Entry Point} &  \textbf{Version} & \textbf{Configuration} \\
        \hline
        Llama 3.1 405b & Meta Platforms & 405 billions & Open-source & Replicate API & meta/meta-llama-3.1-405b-instruct & Temperature: 0.9; Top\_p: 0.6; Top\_k: 40\\ 
        \hline
        Mistral Large & Mistral AI & 123 billions & Open-source & Mistral API & mistral-large-latest & Temperature: 0.3; Top\_p: 1\\ 
        \hline
        GPT-4 & OpenAI, Inc. & Undisclosed & Proprietary & OpenAI API & gpt-4 & Temperature: 1.0; Top\_p: 1.0; \\ 
        \hline
        Claude 3.5 Sonnet & Anthropic PBC & Undisclosed & Proprietary & Anthropic API & claude-3-5-sonnet-20241022 & Temperature: 1.0\\ 
        \hline
    \end{tabular}
        \vspace{-2mm}
    \caption{Description of the selected LLM models.}
    \label{tab:model_descriptions}
\end{table*}

\subsection{LLMs and languages}

We evaluate AI agents using four widely-used and publicly available LLMs, described along with their key details in Table~\ref{tab:model_descriptions}. All models were tested using the default settings recommended by their respective providers. 
For all LLMs, we used the latest available versions at the time of the experiments, conducted from 10 to 15 February 2025.

Our study is conducted in five different written languages: English, French, Arabic, Vietnamese and Mandarin Chinese, to represent a diverse set of linguistic and cultural contexts, covering different scripts, grammatical structures, and global regions. This selection ensures a balanced and comprehensive analysis of language biases across widely spoken and culturally dominant languages.
The template for each game is translated from English into each of the five target languages first by using an automated translator (see details in Supplementary Material Section S3 \cite{suppFairgame}),
and then edited manually  by a native speaker. The personality traits were also revised by native speakers.

\subsection{Games}
\label{sub:games}
We considered two classical game-theoretic scenarios:
\begin{itemize}
    \item \textbf{Prisoner's Dilemma:} Two players face incentives to defect or to cooperate, with mutual cooperation leading to a collectively better payoff. By the theory, the dominant strategy equilibrium results in mutual defection, which is suboptimal for both parties.
    \item \textbf{Battle of the Sexes:} A coordination game involving two players who prefer different end results, but significantly value coordination over disagreement. This scenario highlights the strategic challenge of coordinating on mutually acceptable outcomes despite conflicting individual preferences.
\end{itemize}
We input their description in the template file, using standard game-theoretic narrative \cite{owen2013game} (an example for the Prisoner Dilemma is in Supplementary Material Section S2 \cite{suppFairgame}). 
Each game is associated with a payoff matrix that represents the penalties or rewards incurred by various strategic choices, with player payoffs or gains expressed as negative values of these penalties. The matrices are in the form given by Table \ref{tab:prisoner_conventional}, and is parsed to the \texttt{config} file as described above. 

\begin{table}[ht]
\centering
\begin{tabular}{c|c|c}
& Option A & Option B \\
\hline
Option A & $x_{1,1} = (a_1, a_2 )$ & $x_{1,2} = (b_1, b_2 )$ \\
Option B & $x_{2,1} = (c_1, c_2 )$ & $x_{2,2} = (d_1, d_2 )$ \\
\end{tabular}
    \vspace{-2mm}
\caption{Generic form of the payoff matrix. 
}
    \vspace{-2mm}
\label{tab:prisoner_conventional}
\end{table}
\vspace{-2mm}

To explore strategic variations in adversarial interactions, and assess the sensitivity of each LLM to game parameters, we define multiple configurations of the Prisoner's Dilemma game. Using an established scaling of dilemma strength  \cite{wang2015universal}, fixing other payoff matrix entries, the dilemma strength in the Prisoner's Dilemma decrease with the difference between mutual reward and mutual punishment.
For the  \textit{conventional} configuration with penalties $x_{1,1} = (6, 6)$, $x_{1,2} = (0, 10)$, $x_{2,1} = (10, 0)$ and $x_{2,2} = (2,2)$, this difference is $-2 - (-6) = 4$. 
The \textit{harsh} configuration, with $x_{1,1} = (8, 8)$, $x_{1,2} = (0, 10)$, $x_{2,1} = (10, 0)$ and $x_{2,2} = (5,5)$ and the  \textit{mild} configuration, with $x_{1,1} = (8, 8)$, $x_{1,2} = (0, 10)$, $x_{2,1} = (10, 0)$ and $x_{2,2} = (2,2)$, have differences of $3$ and $6$, respectively.

The Battle of the Sexes uses a single (the most commonly adopted in literature) configuration, with matrix entries (referring to payoffs): $x_{1,1} = (10, 7)$, $x_{1,2} = (0, 0)$, $x_{2,1} = (0, 0)$ and $x_{2,2} = (7,10)$.

\subsection{Set up}
\label{sub:set up}


\begin{figure*}[t]
    \centering
    \includegraphics[width=\textwidth]{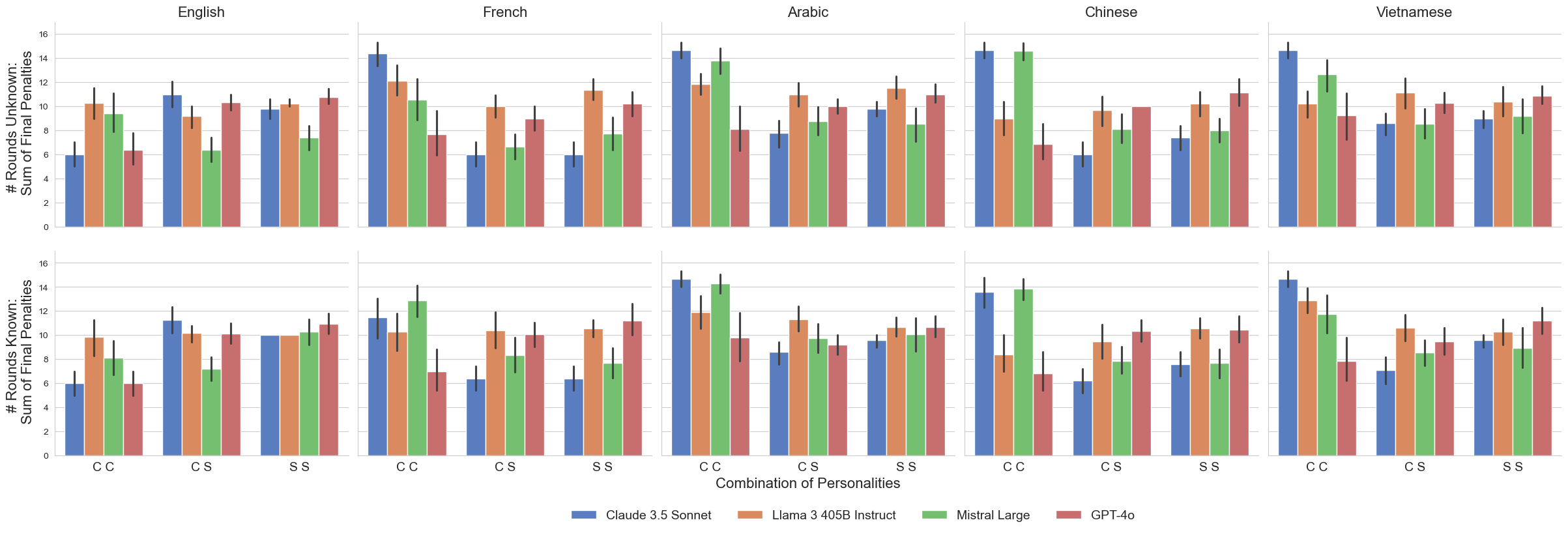}
    \vspace{-5mm}
    \caption{Prisoner's Dilemma: aggregated final scores of the repeated games over repeated experiments, over all three versions, for each LLM, language, combination of personalities and knowledge of opponent's personality. }
    \label{fig:prisoner_all}
\end{figure*}

The experimental configuration employed in this study is as follows. The \texttt{name} of each round depends on the game. The experiments consist of repeated games of 10 rounds each, without earlier stopping condition, for each LLM described in Tab. \ref{tab:model_descriptions}. We test both scenarios where agents are explicitly informed about the total number of rounds, and another where this information is withheld, as this might affect the outcome of the game theoretical predictions \cite{axelrod1981evolution}. Tested languages are: ['en', 'fr', 'ar', 'zh', 'vn'].
We explicitly test the impact of agents' personalities; here, we use 'cooperative' and 'selfish' (future works may even embed the OCEAN framework, or others \cite{hooker2003personality}), to reflect general behavioral attributes commonly utilized and thoroughly documented in game theory literature. Conversely, agent identifiers were intentionally neutral ('agent1' and 'agent2') to eliminate additional variables that could introduce deviations from default behaviors, potentially compromising result interpretability (\textit{cf.} Sec. \ref{sec:interpretation}).
Personality traits are accurately translated into all evaluated languages, whereas agent identifiers remained untranslated, functioning purely as neutral placeholders. Importantly, agents are unaware of their opponent's personality, a condition enforced through setting \textit{opponentPersonalityProb} $= 0$.
All personality permutations were systematically generated, creating scenarios where both agents are cooperative, both selfish, or mixed configurations (one cooperative and one selfish).
The payoffs are tailored for each game type, as described above. No early stopping condition was implemented: all 10 rounds are completed fully. Agents were not permitted to communicate during these experiments, leaving exploration of inter-agent interactions for future research.

The set of all configurations yields 18 distinct games per LLM. Each game is further tested 10 times to ensure statistical reliability. Considering 4 LLM, 5 languages, 10 rounds per game, and 2 decisions per round (one per agent), the experiment generated a total of 72,000 individual decisions.


\section{Results}
\subsection{Prisoner's dilemma}

Fig. \ref{fig:prisoner_all} shows the bar plot (with $95\%$ Confidence Interval) summarizing the test results (in terms of final penalties received by the agents) of the Prisoner's Dilemma games for all three versions (a breakdown for each version, which maintains alike patterns, is reported in Supplementary Material Section S4),
and under two conditions: when agents are unaware of their opponent’s personality and when they are informed. The results are shown across all considered LLMs and languages examined in this study, and for all personality combinations.

In most cases, the preferred end result favors agents defecting, as suggested by game theory. This aligns with the Nash equilibrium of the Prisoner's Dilemma, where mutual defection is the dominant strategy. Nevertheless, there are inconsistencies across languages and combinations of personalities, which suggests that the agents' behavior is influenced by factors beyond the payoff matrix, such as languages and inherent biases present in LLM training data. For instance, penalties are generally lower in English, particularly for GPT-4o and Claude 3.5 Sonnet, when the number of rounds is unknown, indicating more consistent cooperative behavior in their primary training language. 
Broader variability in languages like French or Arabic with high penalties suggests challenges in interpreting the game dynamics due to linguistic or cultural differences, while penalties remain high in Mandarin Chinese and Vietnamese, particularly for Mistral Large.
In mixed personality settings (CS), selfish agents exploit cooperative ones, leading to higher penalties for the latter, consistent with game-theoretic predictions. For selfish pairings (SS), penalties are high but exhibit low variability, as mutual defection is the rational choice. When the number of rounds is unknown, penalties are higher across all settings, reflecting uncertainty that discourages cooperation. Conversely, when rounds are known, penalties decrease, particularly in CC and CS settings, as agents can plan strategies with the endgame in mind.
Claude 3.5 Sonnet and GPT-4o show significant reductions in penalties when rounds are known, especially in English, demonstrating their ability to adapt to game structure. In contrast, Mistral Large shows less sensitivity to this information. Overall, knowing the number of rounds promotes cooperation in CC and CS settings, while SS settings remain unaffected, as defection remains the dominant strategy.


Statistically, we recognize that, when 'selfish' agents are present, there is lower variability in the results, and that some LLMs end up with a broader range of possible outputs than others, also depending on the language used. For instance, all LLM have very narrow distributions in English SS, if the agents know the opponent's personality, but have larger distributions in French, for the same settings. Also, Mistral Large has reduced variability if agents' personalities are known, while GPT-4o has overall larger variability. We will quantitatively measure this variability in Sec. \ref{sub:metrics}. 

\begin{figure}[t]
    \centering
    \includegraphics[width=1\linewidth]{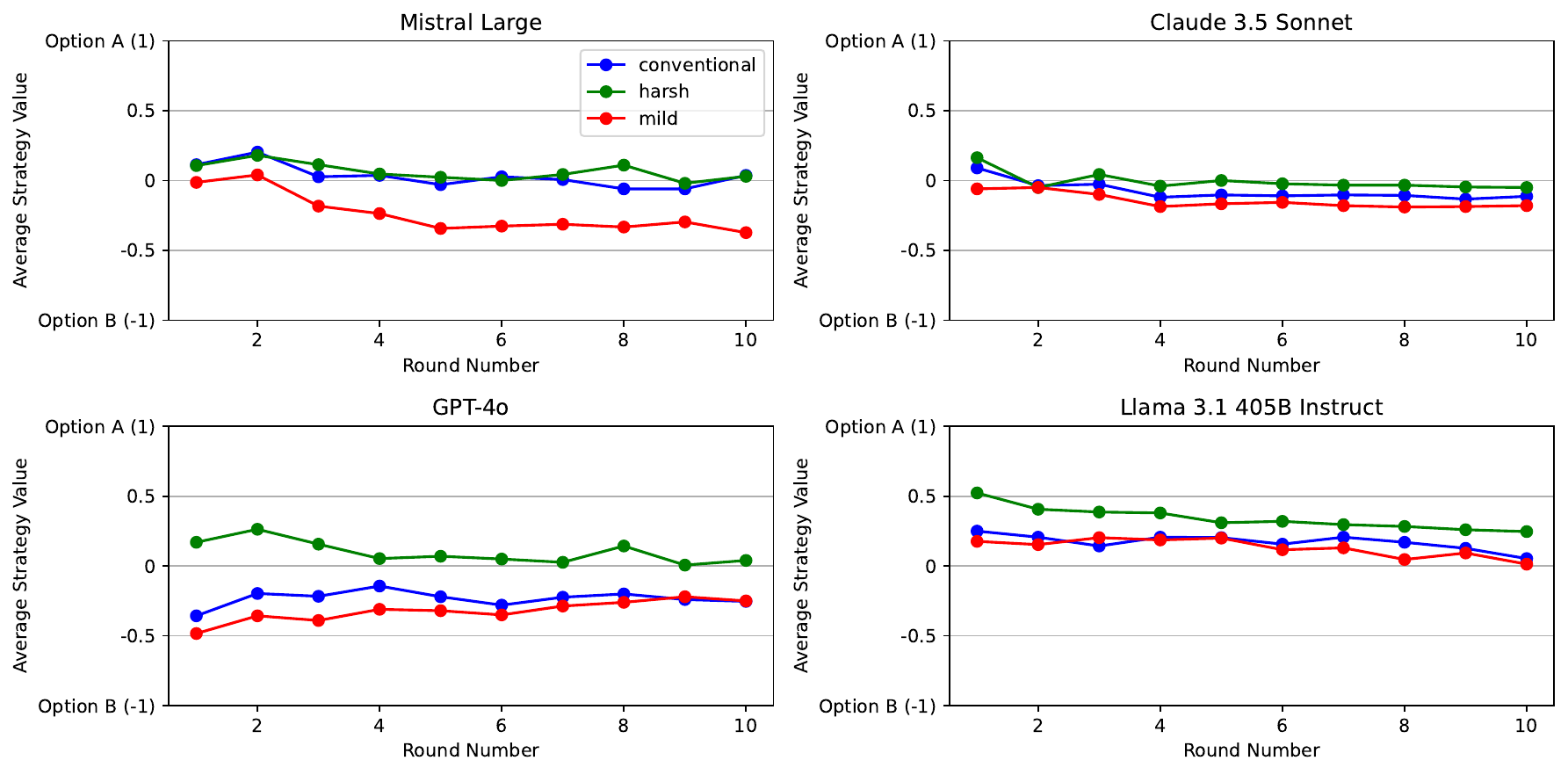}
    \vspace{-5mm}
    \caption{Average trajectory of strategy choices across repeated rounds in all Prisoner’s Dilemma experiments, presented for each LLM and game variant. A value of 1 denotes selection of Option A, which corresponds to defection in this game, while -1 represents Option B (cooperation).}
    \label{fig:prisoner_runs}
\end{figure}

\begin{figure*}
    \centering
    \includegraphics[width=\textwidth]{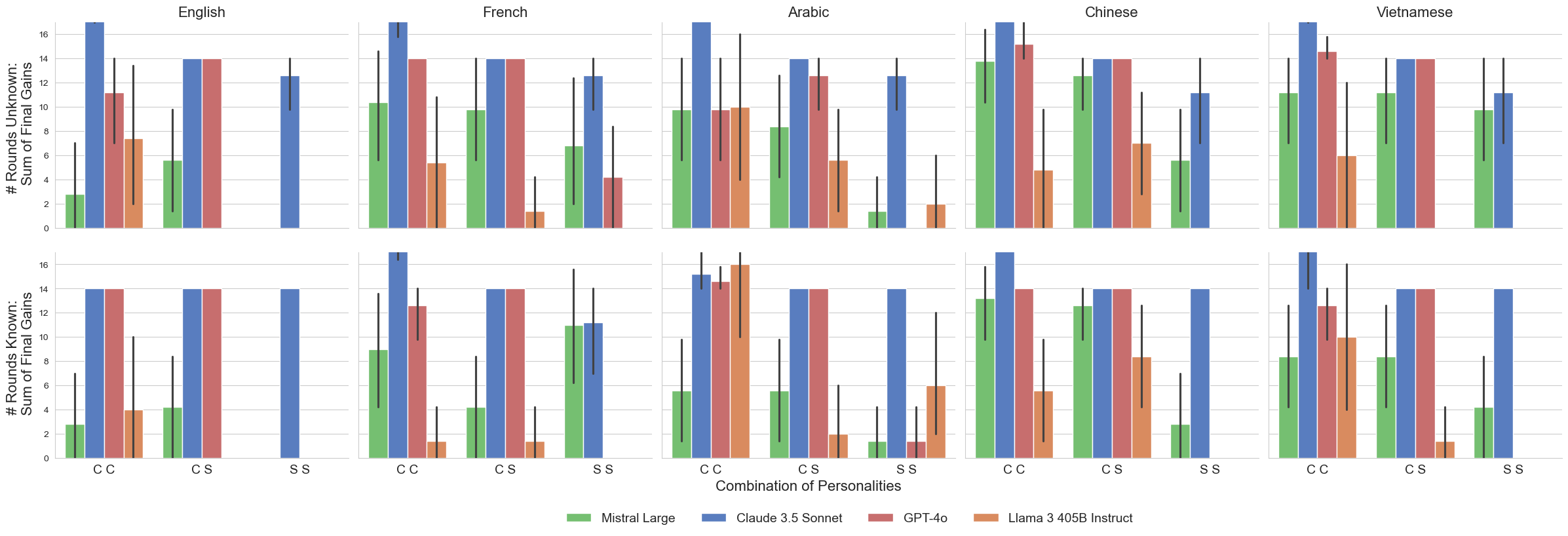}
    \vspace{-5mm}
    \caption{Battle of sexes: aggregated final scores of the repeated games and repeated experiments, over all three versions, for each LLM, language, combination of personalities and knowledge of opponent's personality. Cross language comparison for the conventional configuration.}
    \label{fig:battle_sexes}
\end{figure*}

We also study the evolution of strategies over the rounds. In Fig. \ref{fig:prisoner_runs}, we show the evolution for each version of the game (\textit{cf.} Sec. \ref{sub:games}).
The figure shows that tuning the payoffs changes the strategies adopted during repeated games: all LLMs exhibit, on average, more selfishness under the harsh scenario and more cooperation under the mild scenario compared to the conventional scenario, consistent with the game-theoretic principles of repeated games \cite{wang2015universal}. In the harsh scenario, higher penalties discourage cooperation, leading to more defection. 
Conversely, the mild scenario incentivizes cooperation particularly for Claude 3.5 Sonnet and Llama 3.1 405B. The conventional scenario balances these effects, with strategies stabilizing at intermediate levels. 
We also observe the variability between harsh and mild scenarios, indicative of each model's sensitivity to payoff conditions when making decisions.
Claude 3.5 Sonnet and Llama 3.1 405B demonstrate lower sensitivity, reflected by a narrower spread, while Mistral and GPT-4o show higher sensitivity to the parameters.
Llama 3.1 405B and GPT-4o exhibit less differentiation between the conventional and mild scenarios, whereas Mistral displays smaller differences between harsh and conventional conditions. This behavior suggests distinct decision-making strategies influenced by payoff variations.
Finally, we observe a general downward trend in selfishness over time for Claude 3.5 and Llama 3.1 405B, indicating progressively increasing mutual cooperation among agents, consistent with reciprocal  strategies in repeated games, where agents reciprocate cooperation to maximize long-term payoffs \cite{axelrod1981evolution,wang2015universal}.
Conversely, Mistral Large shows stable cooperation levels under conventional and harsh scenarios but a marked increase in cooperation within the mild scenario.
GPT-4o exhibits divergent patterns, with increasing cooperation in the harsh scenario and increasing selfishness in both mild and conventional scenarios. 
This divergent behavior reflects context-dependent strategic adaptation, potentially due to its higher variability in interpreting payoff structures. These results highlight the interplay between payoff matrices and strategic behavior in repeated games.

\subsection{Battle of sexes}
\label{sub:battle_sexes_results}

Similarly to Fig. \ref{fig:prisoner_all}, Fig. \ref{fig:battle_sexes} presents bar plots (with $95\%$ CI) summarizing the experimental results, over repeated experiments, under two conditions: when agents operate without knowledge of their opponent’s personality and when such information is provided.
The data is reported for all LLM and languages evaluated in this study, and for all combinations of agents' personalities. In this case, Llama 3.1 405B and Mistral Large show the highest internal variability, while Claude Sonnet and GPT-4o are very precise in their final output. Overall, the agents tend to cooperate to maximize the payoffs; however, if the personality if 'selfish', cooperation drops dramatically and low payoffs are achieved. 'French' agents are more cooperative than others, for all LLMs. 
The observed cooperation aligns with the equilibrium in coordination games like the Battle of the Sexes, where agents prioritize coordination over individual preferences to maximize collective payoffs. However, the sharp drop in cooperation with selfish personalities reflects the inherent difficulty in achieving equilibrium when agents prioritize individual payoffs over mutual benefit.

\begin{figure}[t]
    \centering
    \includegraphics[width=0.7\columnwidth]{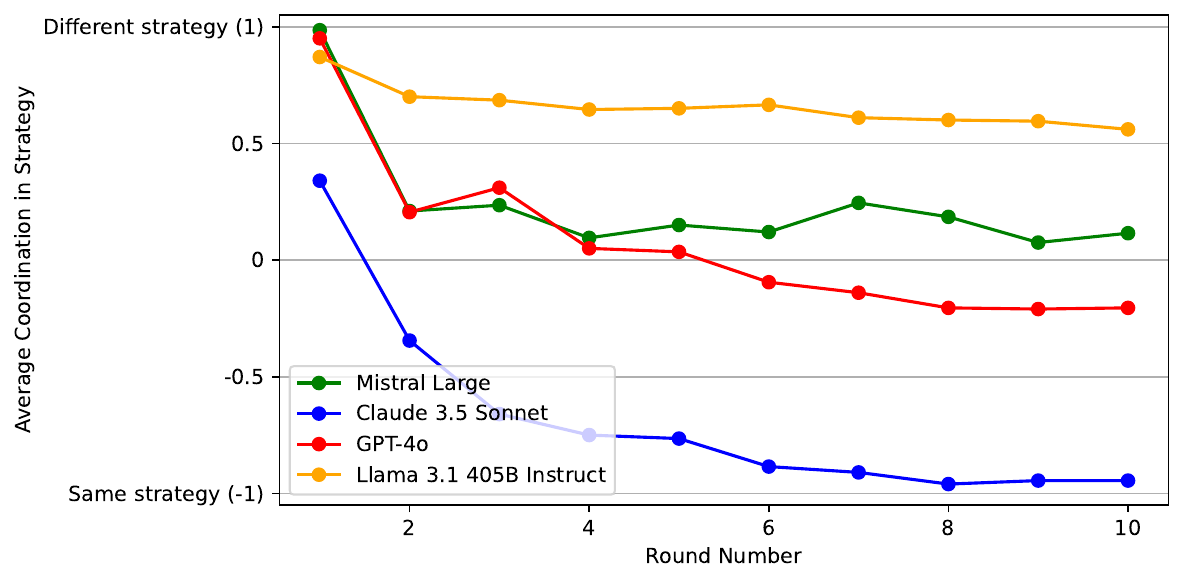}
    \vspace{-3mm}
    \caption{Average evolution of coordination in strategy choices across repeated rounds for all experiments conducted in the Battle of the Sexes, shown separately for each LLM. As this is a coordination game, the plot examines whether the two LLMs select the same option in each round. A value of 1 indicates a mismatch in strategies (one selects Option A, the other Option B), reflecting coordination failure or defective behavior, while -1 indicates alignment in choices, reflecting successful coordination or cooperative behavior.}
    \label{fig:battle_sexes_runs}
\end{figure}

Figure \ref{fig:battle_sexes_runs} shows the evolution of strategy coordination across rounds in the Battle of Sexes game. All LLMs display a gradual improvement in coordination over time. Notably, in the first round, Mistral Large, Llama 3.1 405B Instruct, and GPT-4o tend to select opposing strategies, indicating a lack of coordination, whereas Claude 3.5 Sonnet demonstrates a higher level of alignment from the outset. By the end of the repeated interactions, Llama 3.1 405B Instruct remains largely uncoordinated on average, Mistral Large shows moderate misalignment, GPT-4o achieves moderate coordination, and Claude 3.5 approaches near-perfect coordination.

\subsection{Scoring system}
\label{sub:metrics}

We propose a set of evaluation metrics to score and quantify the key behavioral characteristics of single LLMs and map their tendencies.

\begin{enumerate}
    \item \textbf{Internal Variability ($I_V$)}: the variance of outcomes when the same game scenario is played multiple times, capturing the model’s internal consistency: for an LLM, $I_V = \frac{1}{Z_I}[\text{Var}(\mathbf{y})]$, where $\mathbf{y}$ is the whole results set.
    
    \item \textbf{Cross-Language Inconsistency ($C_I$)}: the standard deviation of results for the same game played in different languages, indicating the instability of the model’s behavior across linguistic contexts: for an LLM, $C_I = \frac{1}{Z_C}[\text{Mean}_{b,c}(\text{Var}_a(\text{Mean}_d(y_{a,b,c,d})))]$, where $a$ indicates languages, $b$ is for personality combinations, $c$ indicates knowledge of rounds, $d$ indicates the rounds $y_{a,b,c,d}$ is the set of results.
    
    \item \textbf{Sensitivity to Payoff ($S_P$)}: the model's responsiveness to changes in incentives. We compute this by measuring the difference in behavior between the \textit{harsh} ($H$) and \textit{mild} ($M$) variants: $S_P = \frac{1}{Z_S}[\text{Mean}_d(|y^{(H)}_d - y^{(M)}_d|)]$, where $y^{(\cdot)}_d$ are the results for each round $d$, averaged over all other features.
    
    \item \textbf{Variability Over Rounds ($V_R$)}: the degree to which the model fluctuates over its strategies, across consecutive rounds of the same game: $V_R=\frac{1}{Z_V}[\text{Mean}_j(\text{Var}_d(y_{d,j}))]$, where $j$ are the game variants and $d$ the rounds.
\end{enumerate}
In all cases, $Z_i = \max[\cdot]$, and are used to normalize the metrics in $[0,1]$. $I_V$, $C_I$, $S_P$ and $V_R$ are then mapped to radar plots, which immediately compare the scores -- and thus the statistical reliability -- of each LLM when addressing a specific game.

\begin{figure}[h]
    \centering
    \includegraphics[width=0.8\columnwidth]{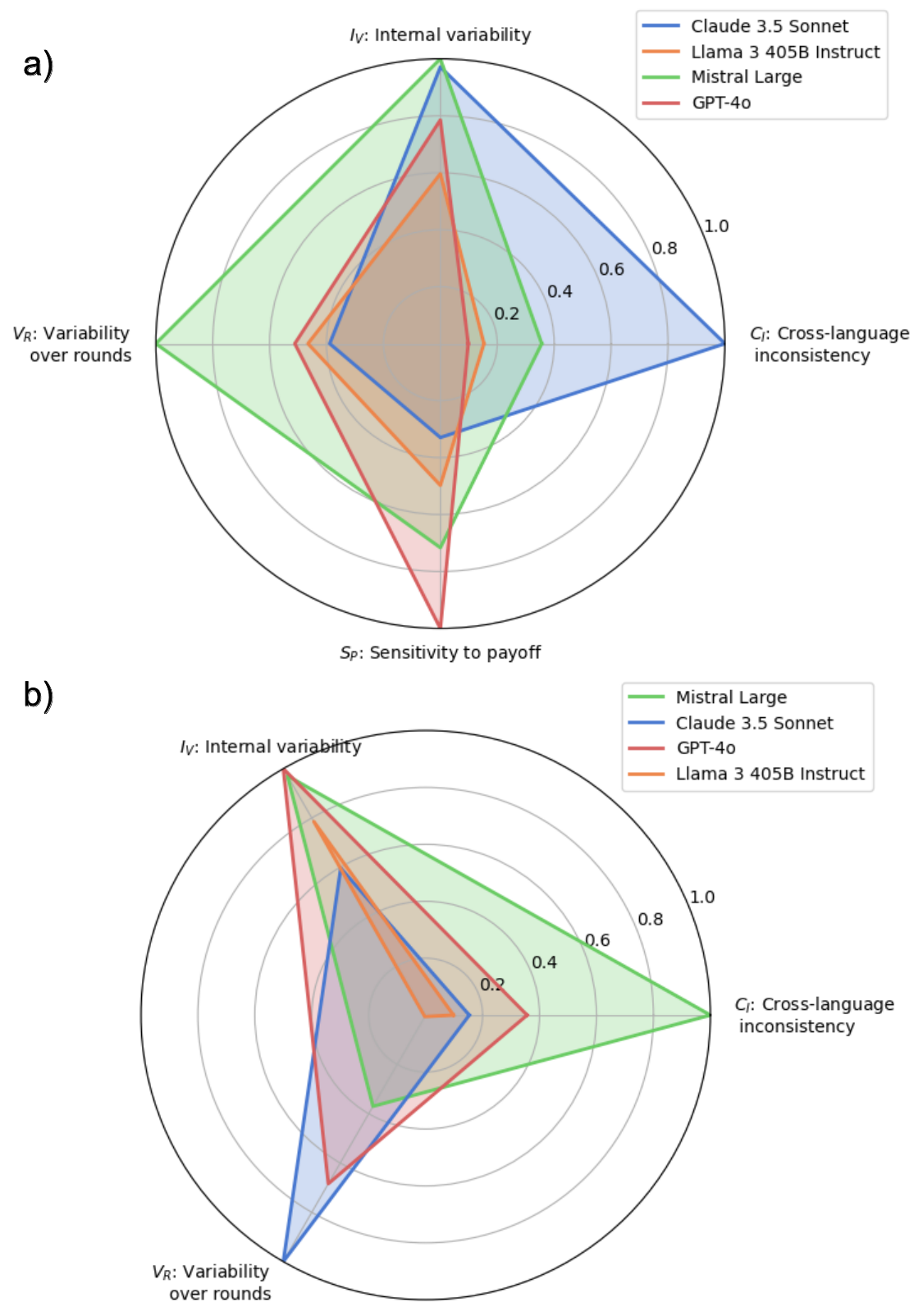}
    \vspace{-2mm}
    \caption{Scoring radar plot for all LLMs over the four dimensions described in Sec. \ref{sub:metrics}, for (a) Prisoner's dilemma; (b) Battle of sexes game.}
    \label{fig:prisoner_spider}
\end{figure}

Fig. \ref{fig:prisoner_spider}a shows such a radar plot for the  Prisoner's dilemma. The higher the metric, the worse the LLM in a certain dimension; the area under the polygon gives immediate information about the overall performance. Mistral Large exhibits the highest variability across evaluation rounds and internal variability, whereas GPT-4o displays the highest sensitivity to payoffs. In contrast, Llama 3.1 405B emerges as the model with the most stable overall behavior across the evaluated dimensions.
Fig. \ref{fig:prisoner_spider}b provides a comparative analysis of the LLMs in the Battle of Sexes game. Given that the LLM were not tested against multiple versions of the game featuring different payoff matrices, the metric $S_P$ was excluded from this evaluation.

The scores computed in Sec. \ref{fig:prisoner_spider} are likely correlated with the degree of influence and bias provided by the training data, as well as with the tendency of LLMs to reduce statistical fluctuations at the cost of not evolving over rounds.
Comparing Figs. \ref{fig:prisoner_spider}a and b reveals that Mistral Large exhibits the greatest inconsistency across different languages, coupled with substantial internal variability, comparable to GPT-4o. Claude 3.5 demonstrates the highest variability across rounds.
Lastly, although Llama 3.1 405B shows notable internal variability, its behavior remains consistent across different languages and rounds; notably, this lower variability for Llama models (coupled with sometimes inconsistent results compared to other LLMs and predictions) was observed in other tasks \cite{buscemi2024chatgpt}, and appears to be a typical trait of the LLM.

\subsection{Interpreting the results}
\label{sec:interpretation}

Thanks to its flexibility and reproducibility, we can use FAIRGAME to test hypothesis about why certain behaviors emerge. For instance, we may hypothesize whether LLMs inherently favor cooperative behaviors over competitive ones, or they possess in-depth knowledge of standard game-theoretical scenarios, including optimal outcomes and effective strategies, which skew their behavior due to influences from training data.

To test the hypothesis, we first asked LLMs about their knowledge level on classic game theory scenarios. Results indicated that LLMs possess substantial familiarity with these games, including the optimal strategies, underlying mechanics, and associated sociological implications contrasting selfish and cooperative human behaviors.

To discern whether observed cooperative behaviors are a result of intrinsic predispositions or pre-existing knowledge, we modified the \texttt{template} file using different narratives for the game. For instance, we reframed the Prisoner's Dilemma into a plane crash scenario, with survivors deciding whether to cooperate in collective hunting tasks in an unhabited island. Instead of numerical payoffs, consequences were communicated qualitatively (e.g., "failure to cooperate results in starvation"). Despite these changes, cooperation levels remained consistently high, in line with Fig. \ref{fig:prisoner_all}. When explicitly queried, LLMs precisely identified these disguised scenarios as variants of the Prisoner's Dilemma, thereby complicating efforts to definitively attribute their behavior to either inherent biases or recognition of known game structures. Finally, we run a second set of experiments introducing distinct identities and personalities to the agents involved in these interactions, so as to assess the impact on behavior. Our findings revealed notable behavioral shifts aligned with the assigned identities. Specifically, pairing archetypal figures such as Adolf Hitler, representing aggressive selfishness, and Mahatma Gandhi, symbolizing peaceful cooperation, resulted in predictable outcomes where the aggressive figure consistently opted for betrayal and the cooperative figure consistently opted to cooperate; this hints to LLMs using prior knowledge on top of the information encoded in the payoff matrices.

\section{Conclusions}

FAIRGAME provides a novel integration of LLMs and game theory, establishing a bidirectional relationship between them. Game theory provides the mathematical foundation for understanding strategic decision-making, interpreting and explaining how AI agents reason and make decisions; LLMs, as experimental tools for data-driven modeling of human decision-making, offer opportunities for empirical validation and exploration of complex interaction scenarios.

Applying game-theoretic approaches to multi-LLM interactions uncovers biases and emergent behaviors, improving interpretability, fairness, efficiency, legal compliance and trust \cite{andras2018trusting, powers2023stuff,buscemi2025llms}. The framework quantifies outcome distributions across games of varied structure and shows that LLMs draw on prior knowledge about the games and their characters -- not merely the pay-off matrices -- when selecting strategies.

Future work can widen the bias spectrum beyond language and personality to nationality, gender, race, age, and more; add communication among players, which might reshape payoffs and strategies \cite{hua2024gametheoreticllmagentworkflow}; and scale beyond two agents, so as to model coalition formation and commitment, as well as group reciprocity \cite{ray2007game,van2012emergence,SONG2025245}. Such studies will clarify how trait combinations shape collective dynamics, guiding team formation, cooperative AI and socially intelligent agents \cite{dafoe2021cooperative,lu2024llms}.

FAIRGAME is readily extendable to incomplete-information, simultaneous-move and sequential games, and to realistic environments where pay-offs are not pre-defined but inferred from the dynamics and outcomes of the actual use case. These scenarios enable questions such as how linguistic or personality biases influence cooperation-defection cycles in evolutionary games, or which incentives stabilize cooperation under uncertainty.


The same tool suits the rising Agentic-AI paradigm, where autonomous systems pursue complex organizational goals with minimal human oversight \cite{acharya2025agentic}. Modeling their interdependencies with FAIRGAME will reveal how diverse cognitive traits and contexts affect performance, supporting configurations that privilege collective benefits, over individual interests, across sectors such as healthcare, finance, manufacturing, autonomous transport, cybersecurity, smart-city infrastructures, and more.


Finally, real-world application of reproducible LLM-game simulations include detecting and mitigating jailbreaking attempts \cite{liu2023jailbreaking}, by framing these interactions as strategic games between an attacking agent and a defensive AI, or developing safer and more performing chatbots for customer assistance or mediation.



\begin{ack}
A.B. is supported by Luxembourg AI Factory.
D.P. is supported by the European Union through the ERC INSPIRE grant (project number 101076926). 
Views and opinions expressed are however those of the authors only and do not necessarily reflect those of the European Union or the European Research Council Executive Agency. 
T.A.H. is supported by EPSRC (grant EP/Y00857X/1).
\end{ack}



\bibliography{references}

@article{wang2015universal,
  title={Universal scaling for the dilemma strength in evolutionary games},
  author={Wang, Zhen and Kokubo, Satoshi and Jusup, Marko and Tanimoto, Jun},
  journal={Phys. Life Rev.},
  volume={14},
  pages={1--30},
  year={2015},
  publisher={Elsevier}
}

@article{dafoe2021cooperative,
  title={Cooperative AI: machines must learn to find common ground},
  author={Dafoe, Allan and Bachrach, Yoram and Hadfield, Gillian and Horvitz, Eric and Larson, Kate and Graepel, Thore},
  journal={Nature},
  volume={593},
  number={7857},
  pages={33--36},
  year={2021},
  publisher={Nature Publishing Group UK London}
}

@book{ray2007game,
  title={A game-theoretic perspective on coalition formation},
  author={Ray, Debraj},
  year={2007},
  publisher={Oxford University Press}
}

@article{van2012emergence,
  title={Emergence of fairness in repeated group interactions},
  author={Van Segbroeck, Sven and Pacheco, Jorge M and Lenaerts, Tom and Santos, Francisco C},
  journal={Phys. Rev. Lett.},
  volume={108},
  number={15},
  pages={158104},
  year={2012},
  publisher={APS}
}

@article{SONG2025245,
title = {On evolution of non-binding commitments},
journal = {Physics of Life Reviews},
volume = {52},
pages = {245-247},
year = {2025},
issn = {1571-0645},
ignoredoi = {https://ignoredoi.org/10.1016/j.plrev.2025.01.006},
author = {Zhao Song and The Anh Han}
}

@article{buscemi2025llms,
  title={{Do LLMs trust AI regulation? Emerging behaviour of game-theoretic LLM agents}},
  author={Buscemi, Alessio and Proverbio, Daniele and Bova, Paolo and Balabanova, Nataliya and Bashir, Adeela and Cimpeanu, Theodor and others},
  journal={arXiv:2504.08640},
  year={2025}
}

@article{liu2023jailbreaking,
  title={Jailbreaking chatgpt via prompt engineering: An empirical study},
  author={Liu, Yi and Deng, Gelei and Xu, Zhengzi and Li, Yuekang and Zheng, Yaowen and Zhang, Ying and others},
  journal={arXiv:2305.13860},
  year={2023}
}

@article{axelrod1981evolution,
  title={The evolution of cooperation},
  author={Axelrod, Robert and Hamilton, William D},
  journal={Science},
  volume={211},
  number={4489},
  pages={1390--1396},
  year={1981},
  publisher={American Association for the Advancement of Science}
}

@article{chen2023utility,
      title={Utility Fairness in Contextual Dynamic Pricing with Demand Learning}, 
      author={Xi Chen and David Simchi-Levi and Yining Wang},
      year={2023},
      journal={arXiv:2311.16528}
}

@article{wang2024large,
  title={Large Language Models Overcome the Machine Penalty When Acting Fairly but Not When Acting Selfishly or Altruistically},
  author={Wang, Zhen and Song, Ruiqi and Shen, Chen and Yin, Shiya and Song, Zhao and Battu, Balaraju and Shi, Lei and Jia, Danyang and Rahwan, Talal and Hu, Shuyue},
  journal={arXiv:2410.03724},
  year={2024}
}

@article{hooker2003personality,
  title={Personality and adult development: Looking beyond the OCEAN},
  author={Hooker, Karen and McAdams, Dan P},
  journal={J. Gerontology B},
  volume={58},
  number={6},
  pages={P311--P312},
  year={2003},
  publisher={Oxford University Press}
}

@article{tessler2024ai,
  title={AI can help humans find common ground in democratic deliberation},
  author={Tessler, Michael Henry and Bakker, Michiel A and Jarrett, Daniel and Sheahan, Hannah and Chadwick, Martin J and others},
  journal={Science},
  volume={386},
  number={6719},
  pages={eadq2852},
  year={2024},
  publisher={American Association for the Advancement of Science}
}

@inproceedings{ramachandran2022contract,
  title={Contract Price Negotiation Using an AI-Based Chatbot},
  author={Ramachandran, Divya and Keshari, Anupam and Tiwari, Manoj Kumar},
  booktitle={Int. Conf. Data An. Pub. Proc. Supply Chain},
  pages={303--310},
  year={2022},
  organization={Springer}
}

@article{buscemi2024newspapers,
  title={Large Language Models’ Detection of Political
Orientation in Newspapers},
  author={Buscemi, Alessio and Proverbio, Daniele},
  journal={arxiv:2406.00018},
  year={2024}
}

@article{buscemi2024chatgpt,
  title={ChatGPT vs Gemini vs LLaMA on Multilingual Sentiment Analysis},
  author={Buscemi, Alessio and Proverbio, Daniele},
  journal={arXiv:2402.01715},
  year={2024}
}

@article{brooks2022artificial,
  title={Artificial bias: the ethical concerns of AI-driven dispute resolution in family matters},
  author={Brooks, Wensdai},
  journal={J. Disp. Resol.},
  pages={117},
  year={2022},
  publisher={HeinOnline}
}

@article{hua2024gametheoreticllmagentworkflow,
      title={Game-theoretic LLM: Agent Workflow for Negotiation Games}, 
      author={Wenyue Hua and Ollie Liu and Lingyao Li and Alfonso Amayuelas and Julie Chen and Lucas Jiang and others},
      year={2024},
      journal={arxiv:2411.05990}, 
}

@misc{githubFairgame,
    author = {{A. Buscemi}},
    year = {2025},
    title = {Fairgame},
    url = {https://github.com/aira-list/FAIRGAME},
}

@misc{suppFairgame,
    author = {{A. Buscemi}},
    year = {2025},
    title = {Fairgame},
    url = {https://github.com/alessio0208/FAIRGAME-ECAI-2025---Supplementary-Material/blob/main/Fairgame_Supplementary_Material.pdf},
}

@book{owen2013game,
  title={Game theory},
  author={Owen, Guillermo},
  year={2013},
  publisher={Emerald Group Publishing}
}

@article{falcao2024making,
  title={Making sense of negotiation and AI: The blossoming of a new collaboration},
  author={Falc{\~a}o Filho, Horacio Arruda},
  journal={Int. J. Commerce Contract.},
  volume={8},
  number={1-2},
  pages={44--64},
  year={2024},
  publisher={SAGE Publications Sage UK: London, England}
}

@article{han2021or,
  title={When to (or not to) trust intelligent machines: Insights from an evolutionary game theory analysis of trust in repeated games},
  author={Han, The Anh and Perret, Cedric and Powers, Simon T},
  journal={Cognitive Sys. Res.},
  volume={68},
  pages={111--124},
  year={2021},
  publisher={Elsevier}
}

@article{HanAICom2022emergent,
  title={Emergent behaviours in multi-agent systems with Evolutionary Game Theory.},
  author={Han, The Anh},
  journal={AI Commun.},
  volume={35},
  number={4},
  year={2022}
}

@article{li2023evaluating,
  title={Evaluating object hallucination in large vision-language models},
  author={Li, Yifan and Du, Yifan and Zhou, Kun and Wang, Jinpeng and Zhao, Wayne Xin and Wen, Ji-Rong},
  journal={arXiv:2305.10355},
  year={2023}
}

@article{li2023halueval,
  title={Halueval: A large-scale hallucination evaluation benchmark for large language models},
  author={Li, Junyi and Cheng, Xiaoxue and Zhao, Wayne Xin and Nie, Jian-Yun and Wen, Ji-Rong},
  journal={arXiv:2305.11747},
  year={2023}
}

@article{balabanova2025media,
  title={Media and responsible AI governance: a game-theoretic and LLM analysis},
  author={Balabanova, Nataliya and Bashir, Adeela and Bova, Paolo and Buscemi, Alessio and Cimpeanu, Theodor and da Fonseca, Henrique Correia and others},
  journal={arXiv:2503.09858},
  year={2025}
}

@article{fulgu2024surprising,
  title={Surprising gender biases in GPT},
  author={Fulgu, Raluca Alexandra and Capraro, Valerio},
  journal={Comp. Human Beha. Rep.},
  volume={16},
  pages={100533},
  year={2024},
  publisher={Elsevier}
}

@article{willis2025will,
  title={Will Systems of LLM Agents Cooperate: An Investigation into a Social Dilemma},
  author={Willis, Richard and Du, Yali and Leibo, Joel Z and Luck, Michael},
  journal={arXiv:2501.16173},
  year={2025}
}

@article{montero2022inferring,
  title={Inferring strategies from observations in long iterated prisoner’s dilemma experiments},
  author={Montero-Porras, Eladio and Gruji{\'c}, Jelena and Fern{\'a}ndez Domingos, Elias and Lenaerts, Tom},
  journal={Sci. Rep.},
  volume={12},
  number={1},
  pages={7589},
  year={2022},
  publisher={Nature Publishing Group UK London}
}

@article{mao2023alympics,
  title={ALYMPICS: LLM Agents Meet Game Theory--Exploring Strategic Decision-Making with AI Agents},
  author={Mao, Shaoguang and Cai, Yuzhe and Xia, Yan and Wu, Wenshan and Wang, Xun and Wang, Fengyi and Ge, Tao and Wei, Furu},
  journal={arXiv:2311.03220},
  year={2023}
}

@article{he2025generative,
  title={Generative ai for game theory-based mobile networking},
  author={He, Long and Sun, Geng and Niyato, Dusit and Du, Hongyang and Mei, Fang and Kang, Jiawen and others},
  journal={IEEE Wireless Commun.},
  volume={32},
  number={1},
  pages={122--130},
  year={2025},
  publisher={IEEE}
}

@article{fontana2024nicer,
  title={Nicer Than Humans: How do Large Language Models Behave in the Prisoner's Dilemma?},
  author={Fontana, Nicol{\'o} and Pierri, Francesco and Aiello, Luca Maria},
  journal={arXiv:2406.13605},
  year={2024}
}

@article{stewart2024distorting,
  title={The distorting effects of producer strategies: Why engagement does not reveal consumer preferences for misinformation},
  author={Stewart, Alexander J and Arechar, Antonio A and Rand, David G and Plotkin, Joshua B},
  journal={Proc. Natl. Acad. Sci.},
  volume={121},
  number={10},
  pages={e2315195121},
  year={2024},
  publisher={National Academy of Sciences}
}

@article{newsham2025personality,
  title={Personality-Driven Decision-Making in LLM-Based Autonomous Agents},
  author={Newsham, Lewis and Prince, Daniel},
  journal={arXiv:2504.00727},
  year={2025}
}

@article{talajic2024strategic,
  title={Strategic Management of Workforce Diversity: An Evolutionary Game Theory Approach as a Foundation for AI-Driven Systems},
  author={Talaji{\'c}, Mirko and Vranki{\'c}, Ilko and Peji{\'c} Bach, Mirjana},
  journal={Information},
  volume={15},
  number={6},
  pages={366},
  year={2024},
  publisher={MDPI}
}

@article{klinkert2024driving,
  title={Driving generative agents with their personality},
  author={Klinkert, Lawrence J and Buongiorno, Stephanie and Clark, Corey},
  journal={arXiv:2402.14879},
  year={2024}
}

@article{he2024afspp,
  title={AFSPP: Agent Framework for Shaping Preference and Personality with Large Language Models},
  author={He, Zihong and Zhang, Changwang},
  journal={arXiv:2401.02870},
  year={2024}
}

@article{newsham2025inducing,
  title={Inducing Personality in LLM-Based Honeypot Agents: Measuring the Effect on Human-Like Agenda Generation},
  author={Newsham, Lewis and Hyland, Ryan and Prince, Daniel},
  journal={arXiv:2503.19752},
  year={2025}
}

@inproceedings{bahtizin2019using,
  title={Using artificial intelligence to optimize intermodal networking of organizational agents within the digital economy},
  author={Bahtizin, AR and Bortalevich, VY and Loginov, EL and Soldatov, Aleksey Ivanovich},
  booktitle={J. Phys: conference series},
  volume={1327},
  pages={012042},
  year={2019},
  organization={IOP Publishing}
}

@article{chaffer2025governing,
  title={Governing the Agent-to-Agent Economy of Trust via Progressive Decentralization},
  author={Chaffer, Tomer Jordi},
  journal={arXiv:2501.16606},
  year={2025}
}

@article{min2010artificial,
  title={Artificial intelligence in supply chain management: theory and applications},
  author={Min, Hokey},
  journal={Int. J. Logistics: Res. Appl.},
  volume={13},
  number={1},
  pages={13--39},
  year={2010},
  publisher={Taylor \& Francis}
}

@article{abaku2024theoretical,
  title={Theoretical approaches to AI in supply chain optimization: Pathways to efficiency and resilience},
  author={Abaku, Emmanuel Adeyemi and Edunjobi, Tolulope Esther and Odimarha, Agnes Clare},
  journal={Int. J. Sci. Tech. Res. Archive},
  volume={6},
  number={1},
  pages={092--107},
  year={2024}
}

@article{ferrara2023fairness,
  title={Fairness and bias in artificial intelligence: A brief survey of sources, impacts, and mitigation strategies},
  author={Ferrara, Emilio},
  journal={Sci},
  volume={6},
  number={1},
  pages={3},
  year={2023},
  publisher={MDPI}
}

@inproceedings{cabrera2023ethical,
  title={Ethical dilemmas, mental health, artificial intelligence, and llm-based chatbots},
  author={Cabrera, Johana and Loyola, M Soledad and Maga{\~n}a, Irene and Rojas, Rodrigo},
  booktitle={Int. Work-Conference Bioinf. Biomed. Eng.},
  pages={313--326},
  year={2023},
  organization={Springer}
}

@article{gichoya2023ai,
  title={AI pitfalls and what not to do: mitigating bias in AI},
  author={Gichoya, Judy Wawira and Thomas, Kaesha and Celi, Leo Anthony and Safdar, Nabile and Banerjee, Imon and Banja, John D and others},
  journal={Brit. J. Radiology},
  volume={96},
  number={1150},
  pages={20230023},
  year={2023},
  publisher={Oxford University Press}
}

@article{andras2018trusting,
  title={Trusting intelligent machines: Deepening trust within socio-technical systems},
  author={Andras, Peter and Esterle, Lukas and Guckert, Michael and Han, The Anh and Lewis, Peter R and Milanovic, Kristina and others},
  journal={IEEE Tech. Soc. Magazine},
  volume={37},
  number={4},
  pages={76--83},
  year={2018},
  publisher={IEEE}
}

@article{hammond2025multi,
  title={Multi-agent risks from advanced ai},
  author={Hammond, Lewis and Chan, Alan and Clifton, Jesse and Hoelscher-Obermaier, Jason and Khan, Akbir and McLean, Euan and Smith, Chandler and Barfuss, Wolfram and Foerster, Jakob and Gaven{\v{c}}iak, Tom{\'a}{\v{s}} and others},
  journal={arXiv:2502.14143},
  year={2025}
}

@article{patel2020leveraging,
  title={Leveraging predictive modeling, machine learning personalization, NLP customer support, and AI chatbots to increase customer loyalty},
  author={Patel, Nikhil and Trivedi, Sandeep},
  journal={Empir. Quests Manage. Essenc.},
  volume={3},
  number={3},
  pages={1--24},
  year={2020}
}

@article{stone2020artificial,
  title={Artificial intelligence (AI) in strategic marketing decision-making: a research agenda},
  author={Stone, Merlin and Aravopoulou, Eleni and Ekinci, Yuksel and Evans, Geraint and Hobbs, Matt and Labib, Ashraf and Laughlin, Paul and Machtynger, Jon and Machtynger, Liz},
  journal={The Bottom Line},
  volume={33},
  number={2},
  pages={183--200},
  year={2020},
  publisher={Emerald Publishing Limited}
}

@article{lu2024llms,
  title={LLMs and generative agent-based models for complex systems research},
  author={Lu, Yikang and Aleta, Alberto and Du, Chunpeng and Shi, Lei and Moreno, Yamir},
  journal={Phys. Life Rev.},
  year={2024},
  publisher={Elsevier}
}

@inproceedings{park2023generative,
  title={Generative agents: Interactive simulacra of human behavior},
  author={Park, Joon Sung and O'Brien, Joseph and Cai, Carrie Jun and Morris, Meredith Ringel and Liang, Percy and Bernstein, Michael S},
  booktitle={Proc. 36th ACM Symp. User Int. Softw. Tech.},
  pages={1--22},
  year={2023}
}

@article{powers2023stuff,
  title        = {{The Stuff We Swim in: Regulation Alone Will Not Lead to Justifiable Trust in AI}},
  author       = {Powers, Simon T and Linnyk, Olena and others},
  year         = 2023,
  journal      = {IEEE Tech. Soc. Mag.},
  publisher    = {Ieee},
  volume       = 42,
  number       = 4,
  pages        = {95--106}
}

@article{el2025towards,
  title={Towards Mechanistic Interpretability of Graph Transformers via Attention Graphs},
  author={El, Batu and Choudhury, Deepro and Li{\`o}, Pietro and Joshi, Chaitanya K},
  journal={arXiv:2502.12352},
  year={2025}
}

@article{ali2025entropy,
  title={Entropy-Lens: The Information Signature of Transformer Computations},
  author={Ali, Riccardo and Caso, Francesco and Irwin, Christopher and Li{\`o}, Pietro},
  journal={arXiv:2502.16570},
  year={2025}
}

@article{acharya2025agentic,
  title={Agentic AI: Autonomous Intelligence for Complex Goals--A Comprehensive Survey},
  author={Acharya, Deepak Bhaskar and Kuppan, Karthigeyan and Divya, B},
  journal={IEEE Access},
  year={2025},
  publisher={IEEE}
}

@misc{buscemi2025strategic,
      title={Strategic Communication and Language Bias in Multi-Agent LLM Coordination}, 
      author={Alessio Buscemi and Daniele Proverbio and Alessandro Di Stefano and The Anh Han and German Castignani and Pietro Liò},
      year={2025},
      eprint={2508.00032},
      archivePrefix={arXiv},
      primaryClass={cs.MA},
      url={https://arxiv.org/abs/2508.00032}, 
}

\end{document}